\documentclass[conference]{IEEEtran}
\IEEEoverridecommandlockouts
\usepackage{cite}
\usepackage{amsmath,amssymb,amsfonts}
\usepackage{algorithmic}
\usepackage{graphicx}
\usepackage{textcomp}
\usepackage{xcolor}
\usepackage{graphicx}
\usepackage{booktabs}
\usepackage{array,multirow,tabularx,ragged2e}
\usepackage{lmodern}
\usepackage{verbatim} 
\def\BibTeX{{\rm B\kern-.05em{\sc i\kern-.025em b}\kern-.08em
    T\kern-.1667em\lower.7ex\hbox{E}\kern-.125emX}}
\graphicspath{{photos/}}
\begin{document}

\title{Image Transformation can make Neural Networks more robust against Adversarial Examples\\
}

\author{\IEEEauthorblockN{Dang Duy Thang }
\IEEEauthorblockA{\textit{Information Security Department } \\
\textit{Institute of Information Security }\\
Yokohama, Japan \\
dgs174101@iisec.ac.jp }
\and
\IEEEauthorblockN{Toshihiro Matsui }
\IEEEauthorblockA{\textit{Information Security Department } \\
\textit{Institute of Information Security }\\
Yokohama, Japan \\
matsui@iisec.ac.jp }
}

\maketitle

\begin{abstract}
Neural networks are being applied in many tasks related to IoT with encouraging results. For example, neural networks can precisely detect human, objects and animal via surveillance camera for security purpose. However, neural networks have been recently found vulnerable to well-designed input samples that called adversarial examples. Such issue causes neural networks to misclassify adversarial examples that are imperceptible to humans. We found giving a rotation to an adversarial example image can defeat the effect of adversarial examples. Using MNIST number images as the original images, we first generated adversarial examples to neural network recognizer, which was completely fooled by the forged examples. Then we rotated the adversarial image and gave them to the recognizer to find the recognizer to regain the correct recognition. Thus, we empirically confirmed rotation to images can protect pattern recognizer based on neural networks from adversarial example attacks.
\end{abstract}

\begin{IEEEkeywords}
Neural networks, security and privacy, adversarial examples
\end{IEEEkeywords}

\section{Introduction}
Recently, neural networks have achieved very impressive success on a wide range of fields like computer vision \cite{szegedy2016rethinking} and natural language processing \cite{sutskever2014sequence}. There are many tasks that have been used by neural networks close to human-performance, such as image classification \cite{he2016deep}, sentence classification \cite{kim2014convolutional}, voice synthesis \cite{van2016wavenet} and object detection \cite{ren2015faster}. In the Internet of Things (IoT), one of the problems is how to reliably process real-world data is captured from IoT devices. And neural networks are considered to be the most promising method to solve this problem \cite{qiu2016survey}. Despite great successes in numerous of applications in IoT \cite{mahdavinejad2017machine}, many machine learning applications are raising great concerns in the field of security and privacy. Recent research has shown that machine learning models are vulnerable to adversarial examples \cite{goodfellowexplaining}. Adversarial examples are well-designed inputs that are created by adding adversarial perturbations. Machine learning systems have been developed following the assumption that the environment is benign during both training and testing. Intuitively, the inputs \(X\) are assumed to all be get from the same distribution at both training and test time. This means that while test inputs \(X\) are news and previously unseen during the training process, they at least have the same properties as the inputs used for training. These assumptions are advantageous for creating a powerful machine learning model but this rule also makes an attacker be able to alter the distribution at the either training time \cite{xiao2018generating} or testing time \cite{biggio2013evasion}. Typical training attacks \cite{huang2011adversarial} try to inject adversarial training data into the original training set to wrongly train the deep learning models. However, most of existing adversarial attacks are focused on testing phase attacks \cite{carlini2017towards}\cite{madry2017towards} because it is more reliable while training phase attacks are more difficult for implementing because attackers should exploit the machine learning system before executing an attack on it. For example, an attacker might slightly modify an image \cite{xiao2018generating} to cause it to be recognized incorrectly or alter the code of an executable file to enable it to bypass a malware detector \cite{grosse2017adversarial}. For dealing with the existence of adversarial samples, many research works are proposing defense mechanisms for adversarial examples. For examples, Papernot et al. \cite{papernot2016distillation} used the distillation algorithm for defending to adversarial perturbations. However, Carnili et al. \cite{carlini2016defensive} pointed out that method is not effective for improving the robustness of a deep neural network system. Weilin et al. \cite{xu2017feature} proposed the feature squeezing for detecting adversarial examples, and there are several other adversarial detection approaches \cite{carlini2017towards}\cite{madry2017towards}.

In this paper, we study the robustness of neural networks through the very simple technique but very effective is rotation. Firstly, we craft the adversarial examples by using the FGSM algorithm \cite{goodfellowexplaining} on the MNIST dataset \cite{lecun2010mnist}. Afterwards, we apply the rotation algorithm on those adversarial examples and evaluate our method through machine learning system \cite{lecun1998gradient}. The results show that our method is very effective for making neural networks more robustness against adversarial examples.

\section{Background and related work}
In this section, we provide the background about adversarial attack and specify some of the notations that we use in this paper. We denote \(x\) is the original image from a given dataset \(X\), and \(y\) denotes the class that is belong to the class space \(Y\). The ground truth label is denoted by \(y_{true}\). And \(x^*\) denotes the adversarial example that is generated from \(x\). Given an input \(x\), its feature vector at layer \(i\) is \(f_i(x)\), and its predicted probability class of \(y\) is \(p(y|x).y_{x}=argmax_{x}p(x|y)\) is predicted class of \(x\). \(L(x,y)\) denotes the loss function of given input \(x\) and target class \(y\).

\subsection{Adversarial attack methods}

The adversarial examples and their counterparts are defined as indistinguishable from humans. Because it is hard to model human perception, researchers use three popular distance metrics to approximate human's perception based on the \(L^p\) norm:
\begin{equation}
||x||_{p}=\bigg(\sum_{i=1}^n|x_{i}|^p\bigg)^\frac{1}{p}
\end{equation}

Researchers usually use \(L^0, L^2, L^\infty\) metrics for expressing the different aspects of visual significance. \(L^0\) counts the number of pixels with different values at corresponding positions in two images. It describes how many pixels are changed between two images. \(L^2\) is used for measuring the Euclidean distance between two images. And \(L^\infty\) will help to measure the maximum difference for all pixels at corresponding positions in two images. There is no agreement which distance metric is the best so it depends on the proposed algorithms.

Szegedy et al. \cite{szegedy2013intriguing} used a method name L-BFGS (Limited-memory Broyden-Fletcher-Goldfarb-Shanno) to create targeted adversarial examples. This method minimize the weighted sum of perturbation size \(\varepsilon \) and loss function \(L(x^*,y_target)\) while constraining the elements of \(x^*\) to be normal pixel value.

Goodfellow et al. \cite{goodfellowexplaining} consumed that adversarial examples can be caused by cumulative effects of high dimensional model weights. They proposed a simple attack method, called Fast Gradient Sign Method (FGSM):
\begin{equation}\label{eq:fgsm}
x^*= x+\varepsilon \cdot sign(\triangledown_xL(x,y))
\end{equation}
where \(\varepsilon \) denotes the perturbation size for crafting adversarial example \(x^*\) from original input \(x\). Given a clean image \(x\), this method try to create a similar image \(x^*\) in \(L^\infty\) neighborhood of \(x\) that fools the target classifier. This leads to maximize loss function \(L(x,y)\) which is the cost of classifying image \(x\) as the target label \(y\). Fast gradient sign method solves this problem by performing one step gradient update from \(x\) in the input space with small size of perturbation \(\varepsilon \). Increasing \(\varepsilon \) will lead to higher and faster attack success rate however it maybe also makes your adversarial sample to be more difference to original input. FGSM computes the gradients for once, so it is much more efficient than L-BFGS. This method is very simple however it is fast and powerful for creating the adversarial examples. So in this paper, we use this method for attack phase. The model is used to create adversarial attacks is called the attacking model. When the attacking model is the target model itself or contains the target model, the resulting attacks are white-box. In this work, we also implement our method on white-box manner. 
\subsection{Defense methods}
Many research works focused on the adversarial training \cite{kurakin2016adversarial1, tramer2017ensemble} to resist  adversarial attacks for a machine learning system from. This strategy aims to use the adversarial examples to train a machine learning model to make it more robust. Some researchers combine data augmentation with adversarial perturbed data for training \cite{szegedy2013intriguing, kurakin2016adversarial1, tramer2017ensemble}. However, this method is more time consuming than traditional training on only clean images, because an extra training dataset will be added to training set and it is clearly that will take more time than in usual. Other defense strategy is pre-processing based methods that try to remove the perturbation noise before feeding data into a machine learning model. Osadchy et al. \cite{osadchy2017no} use some of filters to remove the adversarial noise, such as the median filter, Gaussian low-pass filter. Meng el al. \cite{meng2017magnet} proposed a two phases defense model. First phase is to detect the adversarial input and the second one reforms original input based on the difference between the manifolds of original and adversarial examples. Another adversarial defense direction is based on gradient masking method \cite{tramer2017ensemble}. This defense strategy performs gradient masking typically result in a model that is very smooth in specific directions and neighborhoods of training data, which makes it harder for attackers to find gradients indicating good candidate directions to perturb the input in a damaging way. Papernot et al. \cite{papernot2016distillation} adapts distillation to adversarial defense, and uses the output of another machine learning model as soft labels to train the target model. Nayebi et al. \cite{nayebi2017biologically} use saturating networks for robustness to adversarial examples. In that paper, the loss function is designed to encourage the activations to be in their saturating regime. Gu et al. \cite{gu2014towards} propose the deep contrastive network, which uses a layer-wise contrastive penalty term to achieve output invariance to input perturbation. However, with methods based on gradient masking, attackers can train a substitute model: a copy that imitates the defended model by observing the labels that the defended model assigns to inputs chosen carefully by the adversary.
\subsection{Neural network architecture}
In this section, we will describe about the neural networks (NNs) architecture that we use in this paper. We use Lenet-5 \cite{lecun1998gradient}, that is a convolutional network used in our experiments. NNs learn hierarchical representations of high dimensional inputs used to solve machine learning tasks, including classification, detection or recognition \cite{lecun2015deep}. This network comprises seven layers not counting the input, all of which contain weights (trainable parameters). The input data is a \(32  \times 32\) pixel image. As Fig. 1, convolutional layers are labeled \(Ci\), subsampling layers are labeled \(Si\), and fully connected layers are labeled \(Fi\), where \(i\) is the layer index.

Layer \(C1\) is convolutional layer with six feature maps. Each unit in each feature map is connected to a \(5  \times 5\) neighborhood in the input. The size of the feature maps is \(28  \times 28\) which prevents connection from the input from falling off the boundary. And layer \(C1\) consists 122,304 connections. Layer \(S2\) is a subsampling layer with six feature maps of size \(14\times14\). This layer reduces the size of features in previous layer. Layer \(C3\) is a convolutional layer with 16 feature maps that each unit in each feature map is connected to several \(5\times5\) neighborhoods at identical locations in a subset of \(S2\)'s feature maps. This layer contains 156,000 connections. Next layer is \(S4\), a subsampling layer with 16 feature maps of size \(5\times5\). This layer has 2000 connections. The layer \(C5\) is a convolutional layer that has 120 feature maps. Each unit in this layer is connected to a \(5\times5\) neighborhoods on all 16 of \(S4\)'s feature maps. This layer contains 48,120 connections. The last layer is \(F6\) has 84 units and is fully connected to previous layer \(C5\). This layer consists 10,164 trainable parameters. For more intuitive, the input of each layer \(f_i\) is the output of the previous layer  \(f_{i-1}\) multiplied by a set of weights, which are part of the layer's parameter \(\theta_i\). A neural net can be viewed as a composition of parameterized functions:
\begin{equation}
f: x  \mapsto  f_n ( \theta _n,...,f_2 ( \theta_2,f_1 ( \theta_1,x))...)
\end{equation}
where \(\theta=\left\{\theta_i\right\}\) are parameters learned during training phase. In the case of classification, the network is given a large collection of known input-label pairs \((x,y)\) and adjusts its parameters \(\theta \) to reduce the label prediction error \(f(x)-y\) on these inputs. At test time, the model extrapolates from its training data to make predictions  \(f(x)\) on unseen inputs. For more understanding, the FGSM equation (\ref{eq:fgsm}) that we describe in previous section, can be described as:
\begin{equation}
\delta _x = \varepsilon \cdot sign(\triangledown_xL(f,x,y))
\end{equation}
where \(f\) is the targeted network, \(L(f,x,y)\) is cost function and \(y\) is label of input \(x\). An adversarial sample \(x^*=x+\delta_x\) is successfully crafted when misclassified by convolutional network \(f\) if it satisfies \(f(x^*) \neq f(x)\) while its perturbation factor \(\delta_x\) still remains indistinguishable to humans.

\section{Our system}
The goal of adversarial examples is to make a machine learning model to mis-classify an input data by changing the objective function value based on it's gradients on the adversarial direction.

\subsection{White-box targeted attack}\label{AA}
We consider the white-box targeted attack settings, where the attacker can fully access into the model type, model architecture, all trainable parameters and the adversary aims to change the classifier's prediction to some specific target class. To create adversarial samples that are misclassified by machine learning model, an adversary with knowledge of the model f and its trainable parameters \(\theta\). In this work, we use FGSM \cite{goodfellowexplaining} method for crafting adversarial examples. We define classifier function \(f: \mathbb{R}^n \rightarrow   \begin{Bmatrix}1...k\end{Bmatrix} \) that mapping image pixel value vectors to a particular label. Then we assume that function \(f\) has a loss function \(L: \mathbb{R}^n \times  \begin{Bmatrix}1...k\end{Bmatrix}  \rightarrow  \mathbb{R} \). For an input image \(x \in  \mathbb{R}^n\) and target label \(y \in  \begin{Bmatrix}1...k\end{Bmatrix} \), our system aims to solve the following optimization problem: \(\theta+L(x+\theta,y)\) subject to \(x+\theta \in  \begin{bmatrix}0,1\end{bmatrix}^n \) , where \(\delta\) is perturbation noise that we add to original image \(x\). We have to note that this function method would yield the solution for \(f(x)\) in the case of convex losses, however the neural networks are non-convex so we end up with an approximation in this case.

\subsection{Rotation - affine transformation}
Affine transformations have been widely used in computer vision \cite{jaderberg2015spatial}. So it has an importance role in computer vision. 
Now we define the range of defense that we want to optimize over. For rotation manner that is one of an affine transformations, we find the parameter \((u^*,v^*,\alpha) \) that rotating the adversarial image with degree around the center will make classifier can remove the adversarial noise. Formally, the pixel at position \((u,v)\) is rotated counterclockwise, and then multiplied by a rotation matrix that calculated from the angle \(\alpha \):
\begin{equation}
 \begin{bmatrix}u^*\\v^*\end{bmatrix}= \begin{bmatrix}\cos \alpha &-\sin \alpha \\\sin \alpha &\cos \alpha \end{bmatrix} \cdot  \begin{bmatrix}u\\v\end{bmatrix} 
\end{equation}

So the vectors \( \begin{bmatrix}u\\v\end{bmatrix} \) and \( \begin{bmatrix}u^*\\v^*\end{bmatrix} \) have the same magnitude and they are separated by an given angle \(\alpha \). In our research, we set angle \( \alpha \in \begin{bmatrix}1,90\end{bmatrix}\) and our experiments show that we can find the best angle that it defeats completely the adversarial noise and re-recognize the correct image.

When we apply rotation technique on the adversarial examples that are generated by FGSM, we observe that adversarial examples are failure with rotation. For more intuitive, supposing we have original data  \((x,y)\) and model \(f\). So our prediction label is \(y_p=f(x)\). And the loss function \(L(y,y_p)\) shows us how far \(y_p\) is away from \(y\). When we apply FGSM for crafting adversarial examples, the purpose is to increase the loss function \(L(y,y_p)\) by add a small adversarial noise to original input \(x\). Recall the FGSM algorithm, equation (\ref{eq:fgsm}) will be turn to the equation as bellow:
\begin{equation}\label{eq:rotation}
x^*=x+ \varepsilon \cdot sign \bigg(  \frac{\partial L(y,y_p) }{\partial x} \bigg)
\end{equation} 

We aim to solve equation (\ref{eq:rotation}) by maximize loss function  \(L(y,y_p )\) instead of \(L(x,y)\) in equation (\ref{eq:fgsm}). The logits (vector of raw prediction) is the output of the neural network before we feed them into the softmax activation function for normalizing, it is described as:
\begin{equation}
logits=f(x)
\end{equation}
\begin{equation}y_{pred} = softmax(logits)\end{equation}
\begin{equation}\label{eq:crossentropy}
L(y,y_p) = CrossEntropy(y,y_p)
\end{equation}
By calculating partial derivative of function (\ref{eq:crossentropy}), we have:
\begin{equation}\label{eq:rotation2}
\begin{aligned}
\frac{\partial L(y,y_p) }{\partial x} &=  \frac{\partial L(y,y_p) }{\partial logits} \frac{\partial logits }{\partial x}\\
&= \frac{\partial L(y,y_p) }{\partial f(x)} \frac{\partial f(x) }{\partial x}\\
&= (y_p-y)\frac{\partial f(x)}{\partial x}
\end{aligned}
\end{equation}

\begin{figure}[h]
	\centering
	\includegraphics[width=0.5\textwidth]{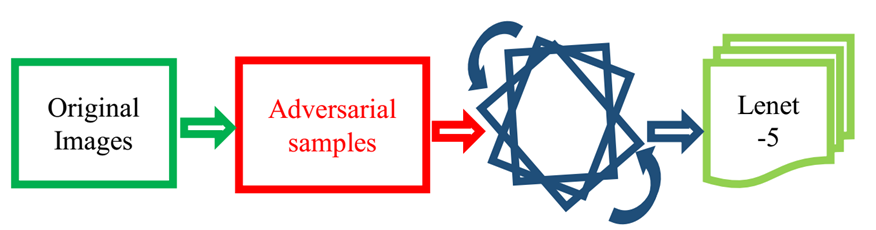}
	\caption{Our system: from original images, we create the adversarial
		examples by using FGSM method on Lenet-5, after that we rotate the
		adversarial images with degree in range [0,90] and finally, using the Lenet-
		5 for classifying}
	\label{fig:our_system}
\end{figure}

From equation (\ref{eq:rotation2}), it is clear that \(\frac{\partial f(x)}{\partial x} \) is influenced by product of trainable weights and activations. For examples, when we have two images with the same label, their activations in any fixed networks are similar and the weights of the network are unchanged. Consequently, \(\frac{\partial f(x)}{\partial x} \) is a constant for any given image \(x\) with the same class. This means the gradient is highly correlated with true label \(y\). Because of this property, when attacker added the adversarial noise, the classifying \(x^*\) becomes a simpler problem than the original problem of classifying \(x\), as \(x^*\) contains extra information from the added noise. However, with a small change in input data (by rotating adversarial images with a particular angle) that system makes an another decision. And in this case, we show that neural network recognizes the rotated adversarial images as true label instead of targeted label. Our system is described as in Fig. \ref{fig:our_system}.
\section{Experiments and results}
Based on the understanding of the problem we described in previous sections, now we can apply our approach to defeat the adversarial examples for protecting a machine learning system.  As our experiments are demonstrated as below, we should note two importance points: a) our approach can remove the adversarial examples, b) our approach still keep neural network's performance is good enough and in some case the neural network performs equal or better than on rotated adversarial images.
\subsection{Experimental setup}
We use the MNIST dataset \cite{lecun2010mnist}, that is a very well-known handwritten digits data in both deep learning and security field. The dataset includes 50,000 training images, 10,000 validation images and 10,000 test images. Each \(28\times28\) gray-scale pixel image is encoded as a vector of intensities whose real values range from 0 (for black color) to 1 (for white color).
\subsection{Results}
We select randomly 10 images in the same category, and name them as original images. We consider the white-box targeted attack so we also randomly choose a targeted label for crafting our targeted adversarial images. In our implementation, the random original input classes are 1 and 6 while random targeted adversarial are 8 and 9 respectively. In attacking phase, we run 20 iterations for crafting adversarial examples with a step size of 0.01 (we choose to take gradient steps in the norm L when using FGSM algorithm).
\begin{figure}[h]
	\centering
	\includegraphics[width=0.4\textwidth]{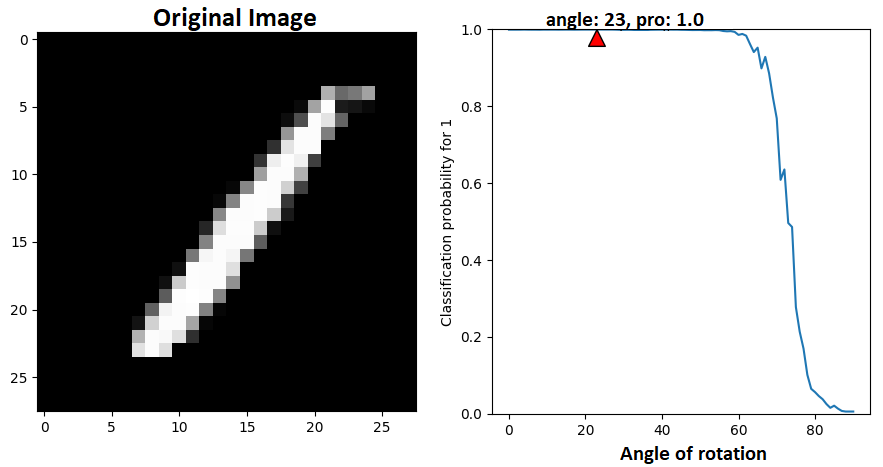}
	\caption{Rotated Original Image (original digit 1, targeted image 8)}
	\label{fig:rot_ori_digit_1}
\end{figure}

\begin{figure}[h]
	\centering
	\includegraphics[width=0.4\textwidth]{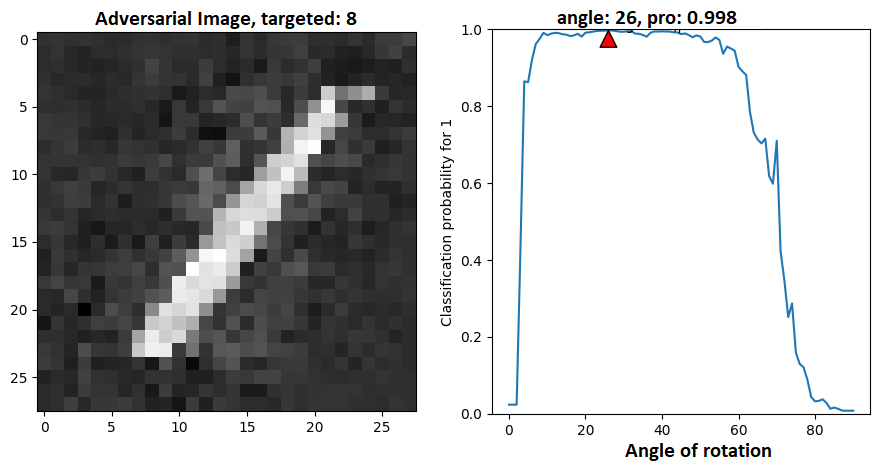}
	\caption{Rotated Adversarial Image (original digit 1, targeted image 8)}
	\label{fig:rot_adv_digit_1}
\end{figure}

In the first implementation with 10 random inputs are digit 1 and targeted adversarial samples is 8. Fig. \ref{fig:rot_adv_digit_1} shows the one adversarial image with targeted digit is 8. The first column shows the adversarial image, and second one describes the classification result after we rotate the adversarial image. The horizontal axis is angle of rotation, that we change angle degree from 0 to 90 and observe the probabilities changing in vertical axis. When degree of angle is 0, it is clear that the classification accuracy of machine learning system recognizes the image as digit 1 is 0. However, this curve rapidly increases with degree of angle and it reaches a peak at angle of rotation is 39 with highest classification rate as digit 1 is around 99.3\%. This result confirms that our approach works well on the adversarial image in this case.

We also compare the difference of classification accuracy between rotated adversarial image and rotated original image. Fig. \ref{fig:rot_ori_digit_1} shows the rotation - affine transformation takes an importance role in image classification problem when it can make classification accuracy lead to the highest score.
\begin{figure}[h]
	\centering
	\includegraphics[width=0.4\textwidth]{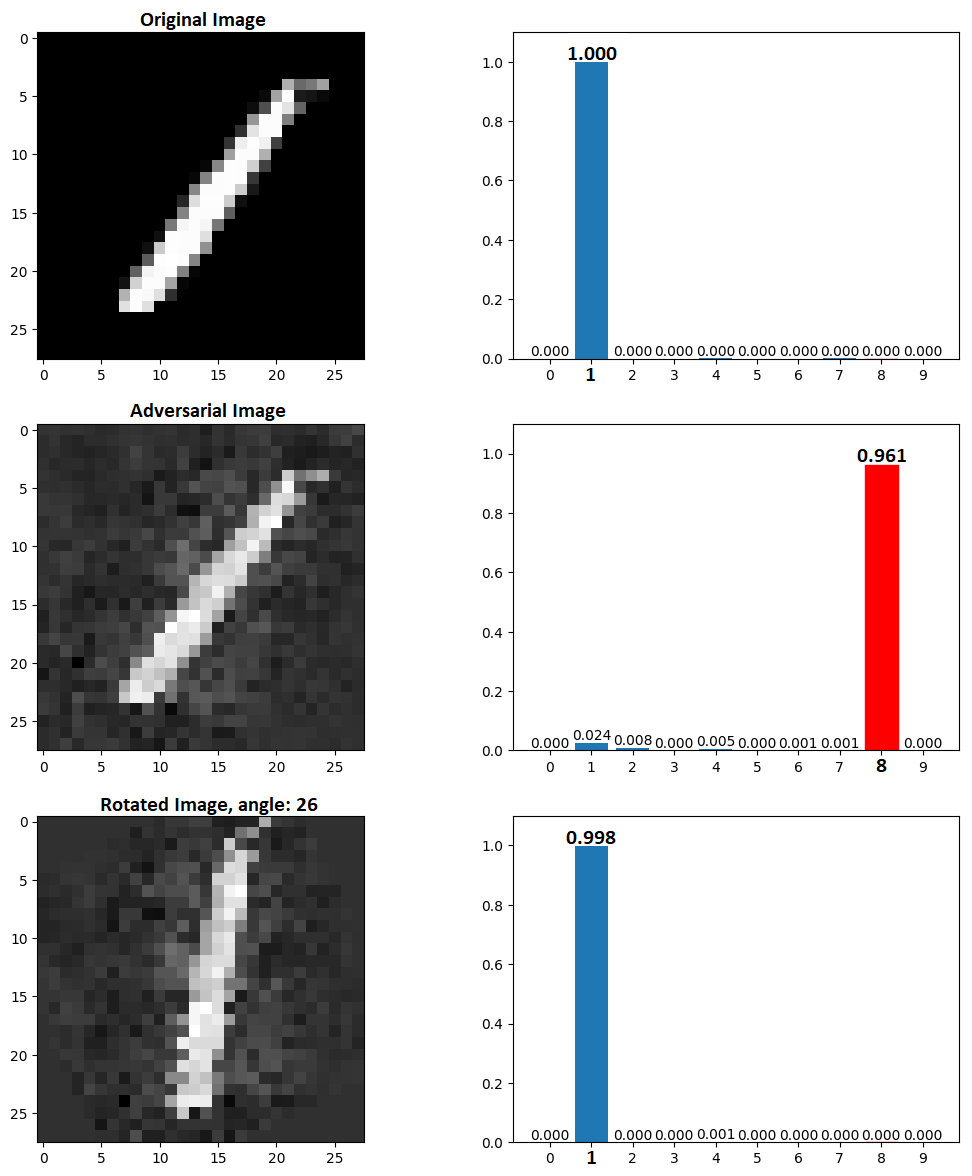}
	\caption{Classification result on original image (digit 1), adversarial
		image (targeted digit 8) and rotated adversarial image}
	\label{fig:gen_adv_digit_1}
\end{figure}

In Fig. \ref{fig:gen_adv_digit_1}, the first row shows the original image (digit 1) and classification accuracy rate for recognizing it as digit 1 is 99.9\% on Lenet-5. The second row shows the adversarial image and classification rate as digit 1 dropped to 0.22\% while machine learning system thinks that image as digit 8 with probability in 97.1\%. And the last row shows the rotated adversarial image at angle of rotation is 39 degree. In this case, the system recognizes it as digit 1 at 99.3\% probability while for digit 8 dropped from 97.1\% to 0.2\%.  So in this implementation, it demonstrates that our approach can completely remove adversarial example and remains the system performance for classification task.
\begin{figure}[h]
	\centering
	\includegraphics[width=0.4\textwidth]{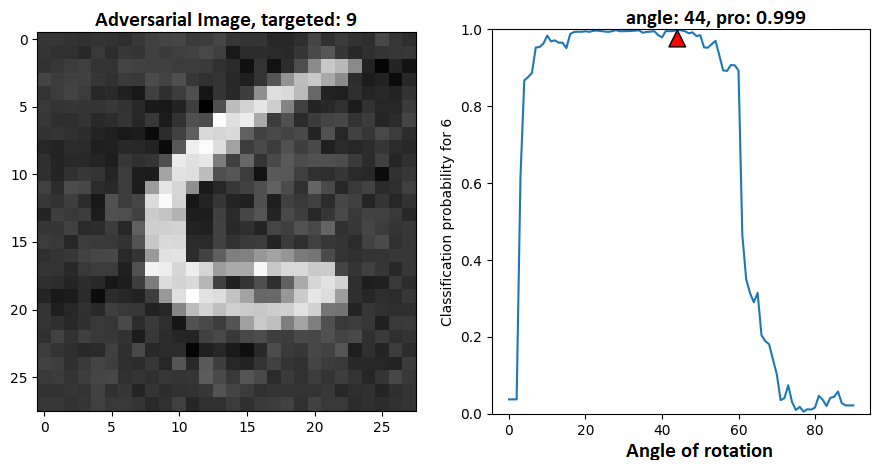}
	\caption{Rotated adversarial image (original digit 6, targeted image digit 9)}
	\label{fig:rot_adv_digit_6}
\end{figure}

We observe the changing classification rate between original image and rotated adversarial image. Table \ref{table:table-1} shows the classification accuracy for original, adversarial and rotated images in our first implementation.
\begin{table}[h!]
	\caption{CLASSIFICATION ACCURACY RATE STATISTICS}
	\label{table:table-1}
	\begin{tabular}{|p{0.8cm}|m{1.5cm}|m{1.5cm}|m{1.5cm}|p{1.2cm}|}
		\hline
		\multirow{2}{*}{\parbox{0.8cm}{Image index}} & \multicolumn{3}{l|}{Classification Accuracy (\%)} & \multirow{3}{*}{\parbox{1.2cm}{Changing rate between (a)-(c)}} \\ \cline{2-4}
		& \begin{tabular}[c]{@{}l@{}}Original\\ digit 1\\ (a)\end{tabular} & \begin{tabular}[c]{@{}l@{}}Adversarial\\ digit 8\\ (b)\end{tabular} & \begin{tabular}[c]{@{}l@{}}Rotated\\ Adv (degree)\\ (c)\end{tabular} &  \\ \hline
		7463 & 100 & 96.1 & 99.8 (26) & 0.2 \\ \hline
		1773 & 94.2 & 96.9 & 96.2 (20) & 2.0 \\ \hline
		9737 & 100 & 96.7 & 99.8 (27) & 0.2 \\ \hline
		7738 & 99.5 & 97.2 & 99.5 (38) & 0 \\ \hline
		9071 & 92.0 & 97.3 & 48.7 (10) & 43.3 \\ \hline
		7399 & 100 & 96.5 & 99.5 (34) & 0.5 \\ \hline
		3765 & 100 & 97.6 & 90.4 (15) & 9.6 \\ \hline
		6670 & 99.9 & 97.5 & 100 (35) & 0.1 \\ \hline
		9896 & 100 & 95.8 & 99.5 (31) & 0.5 \\ \hline
		228 & 100 & 96.0 & 99.4 (19) & 0.6 \\ \hline
	\end{tabular}
\end{table}
The first column in table \ref{table:table-1} is the image index in MNIST dataset that we randomly select for the experiment. We note that when we apply our approach, the neural net can recognize the correct label with slightly decrease accuracy to compare to original recognition. However, in all of cases the adversarial examples are no longer effect on the neural net. 

In the second implementation, we randomly choose 10 inputs as digit 6 and set the targeted adversarial examples is 9. Fig. \ref{fig:rot_adv_digit_6} shows one of 10 adversarial images for demonstration. In this case, the rotated adversarial image is also recognized by system as digit 6 with 100\% probability at angle of rotation is 28 degree.
\begin{figure}[h]
	\centering
	\includegraphics[width=0.35\textwidth]{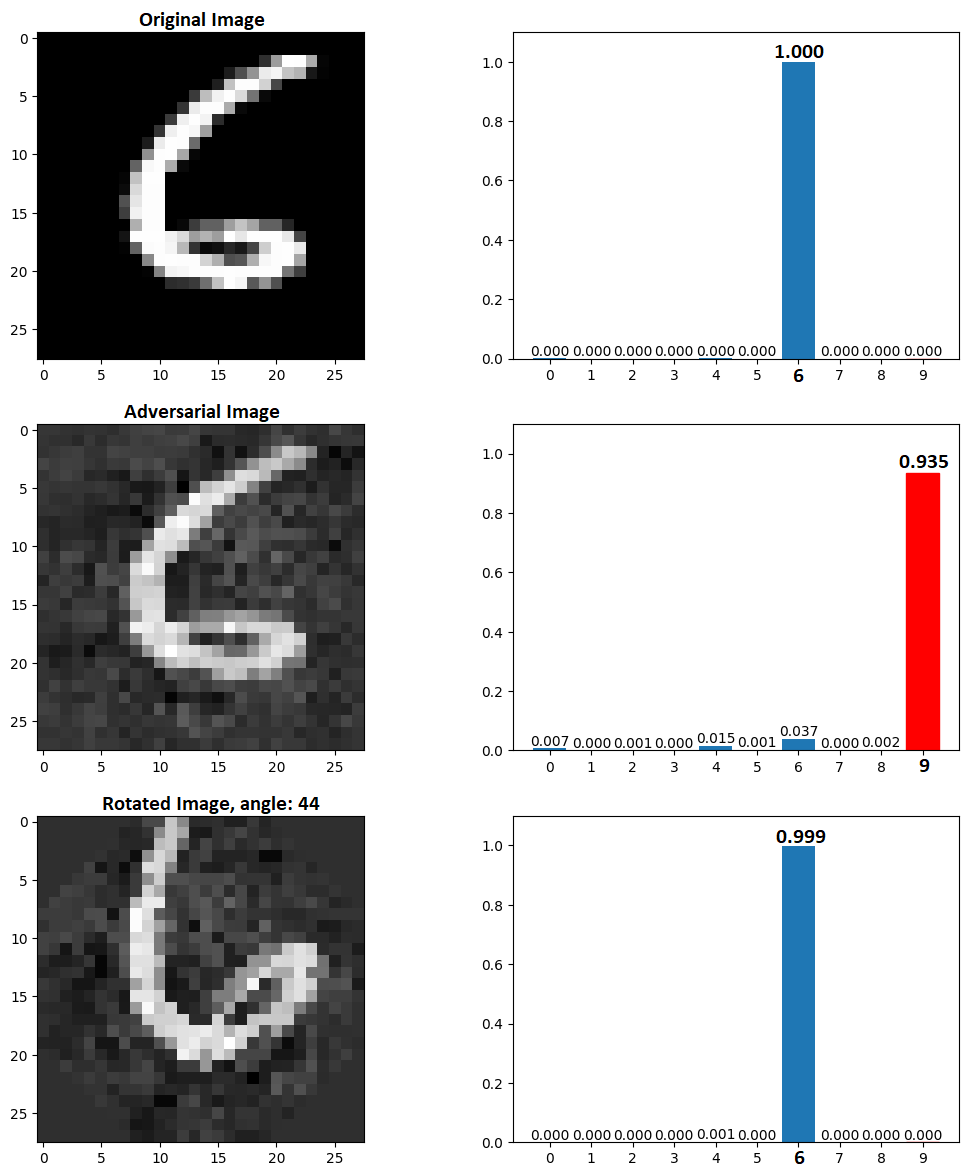}
	\caption{Classification result on original image (digit 6), adversarial
		image (targeted digit is 9) and rotated adversarial image}
	\label{fig:gen_adv_digit_6}
\end{figure}

In Fig. \ref{fig:gen_adv_digit_6}, for given input image digit 6, the system recognizes it as digit 6 with 100\% confidence. In attacking phase, this classification rate dropped to 0.21\% for recognizing digit 6 and 95.6\% for targeted digit 9. Surprisingly, in this case after rotating, the classification rate as digit 6 recovers in a perfect rate at 100\%. So that means our approach completely removes the influence of adversarial and keeps the system up to high performance. 

\section{Conclusion}
Adversarial attack so far is a very serious problem for security and privacy on machine learning system. Our research work provide evidence that the neural networks can be made more robustness to adversarial attacks. As our theory and experiments, we now can develop a powerful defense method for adversarial problem. Our experiments on MNIST have not reached the best on all of cases. However, our results already show that our approach lead to significant increase in the robustness of the neural network. And we also believe that our findings will be further explored in our future work.
\section*{Acknowledgment}

We would like to thank Professor Akira Otsuka for his helpful and valuable comments. This work is supported by Iwasaki Tomomi Scholarship.

\bibliographystyle{IEEEtran}
\bibliography{IEEEabrv,mybibfile}

\begin{thebibliography}{10}
\providecommand{\url}[1]{#1}
\csname url@samestyle\endcsname
\providecommand{\newblock}{\relax}
\providecommand{\bibinfo}[2]{#2}
\providecommand{\BIBentrySTDinterwordspacing}{\spaceskip=0pt\relax}
\providecommand{\BIBentryALTinterwordstretchfactor}{4}
\providecommand{\BIBentryALTinterwordspacing}{\spaceskip=\fontdimen2\font plus
\BIBentryALTinterwordstretchfactor\fontdimen3\font minus
  \fontdimen4\font\relax}
\providecommand{\BIBforeignlanguage}[2]{{%
\expandafter\ifx\csname l@#1\endcsname\relax
\typeout{** WARNING: IEEEtran.bst: No hyphenation pattern has been}%
\typeout{** loaded for the language `#1'. Using the pattern for}%
\typeout{** the default language instead.}%
\else
\language=\csname l@#1\endcsname
\fi
#2}}
\providecommand{\BIBdecl}{\relax}
\BIBdecl

\bibitem{szegedy2016rethinking}
C.~Szegedy, V.~Vanhoucke, S.~Ioffe, J.~Shlens, and Z.~Wojna, ``Rethinking the
  inception architecture for computer vision,'' in \emph{Proceedings of the
  IEEE conference on computer vision and pattern recognition}, 2016, pp.
  2818--2826.

\bibitem{sutskever2014sequence}
I.~Sutskever, O.~Vinyals, and Q.~V. Le, ``Sequence to sequence learning with
  neural networks,'' in \emph{Advances in neural information processing
  systems}, 2014, pp. 3104--3112.

\bibitem{he2016deep}
K.~He, X.~Zhang, S.~Ren, and J.~Sun, ``Deep residual learning for image
  recognition,'' in \emph{Proceedings of the IEEE conference on computer vision
  and pattern recognition}, 2016, pp. 770--778.

\bibitem{kim2014convolutional}
Y.~Kim, ``Convolutional neural networks for sentence classification,''
  \emph{arXiv preprint arXiv:1408.5882}, 2014.

\bibitem{van2016wavenet}
A.~Van Den~Oord, S.~Dieleman, H.~Zen, K.~Simonyan, O.~Vinyals, A.~Graves,
  N.~Kalchbrenner, A.~W. Senior, and K.~Kavukcuoglu, ``Wavenet: A generative
  model for raw audio.'' in \emph{SSW}, 2016, p. 125.

\bibitem{ren2015faster}
S.~Ren, K.~He, R.~Girshick, and J.~Sun, ``Faster r-cnn: Towards real-time
  object detection with region proposal networks,'' in \emph{Advances in neural
  information processing systems}, 2015, pp. 91--99.

\bibitem{qiu2016survey}
J.~Qiu, Q.~Wu, G.~Ding, Y.~Xu, and S.~Feng, ``A survey of machine learning for
  big data processing,'' \emph{EURASIP Journal on Advances in Signal
  Processing}, vol. 2016, no.~1, p.~67, 2016.

\bibitem{erfani2016high}
S.~M. Erfani, S.~Rajasegarar, S.~Karunasekera, and C.~Leckie,
  ``High-dimensional and large-scale anomaly detection using a linear one-class
  svm with deep learning,'' \emph{Pattern Recognition}, vol.~58, pp. 121--134,
  2016.

\bibitem{mahdavinejad2017machine}
M.~S. Mahdavinejad, M.~Rezvan, M.~Barekatain, P.~Adibi, P.~Barnaghi, and A.~P.
  Sheth, ``Machine learning for internet of things data analysis: A survey,''
  \emph{Digital Communications and Networks}, 2017.

\bibitem{goodfellowexplaining}
I.~J. Goodfellow, J.~Shlens, and C.~Szegedy, ``Explaining and harnessing
  adversarial examples,'' \emph{arXiv preprint arXiv:1412.6572}, 2015.

\bibitem{xiao2018generating}
C.~Xiao, B.~Li, J.-Y. Zhu, W.~He, M.~Liu, and D.~Song, ``Generating adversarial
  examples with adversarial networks,'' \emph{arXiv preprint arXiv:1801.02610},
  2018.

\bibitem{biggio2013evasion}
B.~Biggio, I.~Corona, D.~Maiorca, B.~Nelson, N.~{\v{S}}rndi{\'c}, P.~Laskov,
  G.~Giacinto, and F.~Roli, ``Evasion attacks against machine learning at test
  time,'' in \emph{Joint European conference on machine learning and knowledge
  discovery in databases}.\hskip 1em plus 0.5em minus 0.4em\relax Springer,
  2013, pp. 387--402.

\bibitem{huang2011adversarial}
L.~Huang, A.~D. Joseph, B.~Nelson, B.~I. Rubinstein, and J.~Tygar,
  ``Adversarial machine learning,'' in \emph{Proceedings of the 4th ACM
  workshop on Security and artificial intelligence}.\hskip 1em plus 0.5em minus
  0.4em\relax ACM, 2011, pp. 43--58.

\bibitem{carlini2017towards}
N.~Carlini and D.~Wagner, ``Towards evaluating the robustness of neural
  networks,'' in \emph{2017 IEEE Symposium on Security and Privacy (SP)}.\hskip
  1em plus 0.5em minus 0.4em\relax IEEE, 2017, pp. 39--57.

\bibitem{kurakin2016adversarial}
A.~Kurakin, I.~Goodfellow, and S.~Bengio, ``Adversarial examples in the
  physical world,'' \emph{arXiv preprint arXiv:1607.02533}, 2016.

\bibitem{madry2017towards}
A.~Madry, A.~Makelov, L.~Schmidt, D.~Tsipras, and A.~Vladu, ``Towards deep
  learning models resistant to adversarial attacks,'' \emph{arXiv preprint
  arXiv:1706.06083}, 2017.

\bibitem{grosse2017adversarial}
K.~Grosse, N.~Papernot, P.~Manoharan, M.~Backes, and P.~McDaniel, ``Adversarial
  examples for malware detection,'' in \emph{European Symposium on Research in
  Computer Security}.\hskip 1em plus 0.5em minus 0.4em\relax Springer, 2017,
  pp. 62--79.

\bibitem{papernot2016distillation}
N.~Papernot, P.~McDaniel, X.~Wu, S.~Jha, and A.~Swami, ``Distillation as a
  defense to adversarial perturbations against deep neural networks,'' in
  \emph{2016 IEEE Symposium on Security and Privacy (SP)}.\hskip 1em plus 0.5em
  minus 0.4em\relax IEEE, 2016, pp. 582--597.

\bibitem{carlini2016defensive}
N.~Carlini and D.~Wagner, ``Defensive distillation is not robust to adversarial
  examples,'' \emph{arXiv preprint arXiv:1607.04311}, 2016.

\bibitem{xu2017feature}
W.~Xu, D.~Evans, and Y.~Qi, ``Feature squeezing: Detecting adversarial examples
  in deep neural networks,'' \emph{arXiv preprint arXiv:1704.01155}, 2017.

\bibitem{lecun2010mnist}
Y.~LeCun, C.~Cortes, and C.~Burges, ``Mnist handwritten digit database,''
  \emph{AT\&T Labs [Online]. Available: http://yann. lecun. com/exdb/mnist},
  vol.~2, 2010.

\bibitem{lecun1998gradient}
Y.~LeCun, L.~Bottou, Y.~Bengio, and P.~Haffner, ``Gradient-based learning
  applied to document recognition,'' \emph{Proceedings of the IEEE}, vol.~86,
  no.~11, pp. 2278--2324, 1998.

\bibitem{szegedy2013intriguing}
C.~Szegedy, W.~Zaremba, I.~Sutskever, J.~Bruna, D.~Erhan, I.~Goodfellow, and
  R.~Fergus, ``Intriguing properties of neural networks,'' \emph{arXiv preprint
  arXiv:1312.6199}, 2013.

\bibitem{kurakin2016adversarial1}
A.~Kurakin, I.~Goodfellow, and S.~Bengio, ``Adversarial machine learning at
  scale,'' \emph{arXiv preprint arXiv:1611.01236}, 2016.

\bibitem{tramer2017ensemble}
F.~Tram{\`e}r, A.~Kurakin, N.~Papernot, I.~Goodfellow, D.~Boneh, and
  P.~McDaniel, ``Ensemble adversarial training: Attacks and defenses,''
  \emph{arXiv preprint arXiv:1705.07204}, 2017.

\bibitem{osadchy2017no}
M.~Osadchy, J.~Hernandez-Castro, S.~Gibson, O.~Dunkelman, and
  D.~P{\'e}rez-Cabo, ``No bot expects the deepcaptcha! introducing immutable
  adversarial examples, with applications to captcha generation,'' \emph{IEEE
  Transactions on Information Forensics and Security}, vol.~12, no.~11, pp.
  2640--2653, 2017.

\bibitem{meng2017magnet}
D.~Meng and H.~Chen, ``Magnet: a two-pronged defense against adversarial
  examples,'' in \emph{Proceedings of the 2017 ACM SIGSAC Conference on
  Computer and Communications Security}.\hskip 1em plus 0.5em minus 0.4em\relax
  ACM, 2017, pp. 135--147.

\bibitem{papernot2017practical}
N.~Papernot, P.~McDaniel, I.~Goodfellow, S.~Jha, Z.~B. Celik, and A.~Swami,
  ``Practical black-box attacks against machine learning,'' in
  \emph{Proceedings of the 2017 ACM on Asia Conference on Computer and
  Communications Security}.\hskip 1em plus 0.5em minus 0.4em\relax ACM, 2017,
  pp. 506--519.

\bibitem{nayebi2017biologically}
A.~Nayebi and S.~Ganguli, ``Biologically inspired protection of deep networks
  from adversarial attacks,'' \emph{arXiv preprint arXiv:1703.09202}, 2017.

\bibitem{gu2014towards}
S.~Gu and L.~Rigazio, ``Towards deep neural network architectures robust to
  adversarial examples,'' \emph{arXiv preprint arXiv:1412.5068}, 2014.

\bibitem{lecun2015deep}
Y.~LeCun, Y.~Bengio, and G.~Hinton, ``Deep learning,'' \emph{nature}, vol. 521,
  no. 7553, p. 436, 2015.

\bibitem{jaderberg2015spatial}
M.~Jaderberg, K.~Simonyan, A.~Zisserman \emph{et~al.}, ``Spatial transformer
  networks,'' in \emph{Advances in neural information processing systems},
  2015, pp. 2017--2025.

\end{thebibliography}
\end{document}